\documentclass[10pt,twocolumn,letterpaper]{article}

\usepackage{iccv}
\usepackage{times}
\usepackage{epsfig}
\usepackage{graphicx}
\usepackage{amsmath}
\usepackage{amssymb}
\usepackage{subfigure}

\usepackage[breaklinks=true,bookmarks=false]{hyperref}

\iccvfinalcopy 

\def\iccvPaperID{****} 

\begin{document}

\title{Discriminative-Generative Representation Learning for One-Class Anomaly Detection}

\author{
Xuan Xia$^1$\\ {\tt\small {xiaxuan@cuhk.edu.cn}}
\and
Xizhou Pan$^1$\\ {\tt\small {panxizhou@cuhk.edu.cn}}
\and
Xing He$^1$\\ {\tt\small {hexing@cuhk.edu.cn}}
\and
Jingfei Zhang$^1$\\ {\tt\small {117010378@cuhk.edu.cn}}
\and
Ning Ding$^{1,2*}$\\ {\tt\small {dingning@cuhk.edu.cn}}
\and
Lin Ma$^{1*}$\\ {\tt\small {malin@cuhk.edu.cn}}
\and
$^1${\small Shenzhen Institute of Artificial Intelligence and Robotics for Society, Shenzhen 518129, P. R. China.}\\
$^2${\small Institute of Robotics and Intelligent Manufacturing, Chinese University of Hong Kong, 
Shenzhen, 518172, P. R. China.}\\
$^*${\small Corresponding author.}
}

\maketitle

\ificcvfinal\thispagestyle{plain}\fi

\begin{abstract}
As a kind of generative self-supervised learning methods, generative adversarial nets have been widely studied in the field of anomaly detection. However, the representation learning ability of the generator is limited since it pays too much attention to pixel-level details, and generator is difficult to learn abstract semantic representations from label prediction pretext tasks as effective as discriminator. In order to improve the representation learning ability of generator, we propose a self-supervised learning framework combining generative methods and discriminative methods. The generator no longer learns representation by reconstruction error, but the guidance of discriminator, and could benefit from pretext tasks designed for discriminative methods. Our discriminative-generative representation learning method has performance close to discriminative methods and has a great advantage in speed. Our method used in one-class anomaly detection task significantly outperforms several state-of-the-arts on multiple benchmark data sets, increases the performance of the top-performing GAN-based baseline by 6\% on CIFAR-10 and 2\% on MVTAD. What’s more, ablation studies show that absolute position information deteriorates representational learning ability of generative methods in geometric transformation tasks, and has different effects on the representational learning ability of discriminative methods in different geometric transformation tasks, which provides a criterion for the use of position information.
\end{abstract}


\section{Introduction}
Anomalies could be errors in the data or sometimes are previously unknown out-of-distribution samples. As a classic pattern recognition task, anomaly detection has made great progress in the era of deep learning. However, one-class anomaly detection (OCAD) in unsupervised scenarios is still a challenging task due to the difficulty of model establishment and the absence of labels. 

OCAD aims to identify patterns that do not belong to the normal data distribution \cite{chandola2009anomaly}. A typical solution is to map the normal data to a definite distribution in a latent space, thus identifying the data outside the distribution as anomalies. Among many methods, generative adversarial nets (GAN) has been widely used due to its advantages in distribution fitting. At the beginning, researchers usually use auto-encoder structure GAN for data encoding and reconstruction \cite{akcay2018ganomaly,pidhorskyi2018generative,perera2019ocgan}, and measure anomalies by reconstruction error. However, studies on representation learning show that generative learning methods have inherent defects in feature learning. Most GAN-based methods don’t have strong enough constraints on the learning of representations, which is crucial to distinguish abnormal from normal. Pixel-wise reconstruction may result in the loss of important semantic information, and leads to the degradation of anomaly detection performance.

\begin{figure*}[ht]
\centerline{\includegraphics[width=1\textwidth]{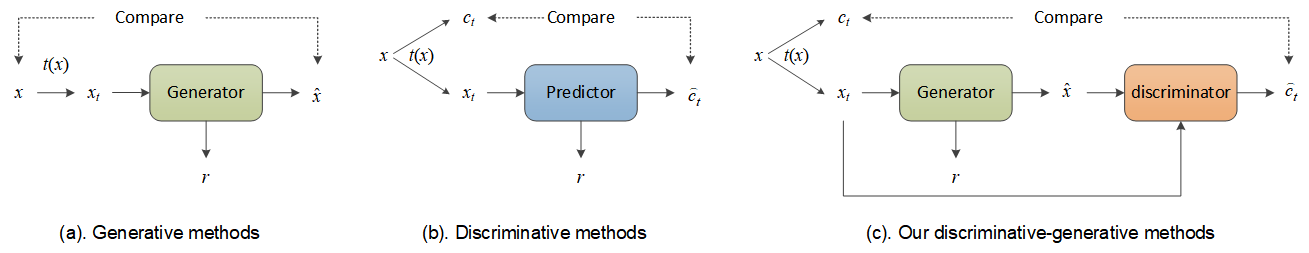}}
\caption{The frameworks of generative methods, discriminative methods and our discriminative-generative methods in self-supervised learning. (a) Generative methods learn representations $r$ by the reconstruction of input $x$, (b) discriminative methods learn representations $r$ by the prediction of label $c$. (c) We reuse the discriminator as a predictor to guide the generator to generate $x$ that match the pretext task labels $c_t$, thus the generator can learn representation $r$ by using pretext tasks designed for discriminative methods.}
\label{Figure1}
\end{figure*}

Therefore, some scholars begin to explore how to learn features efficiently from the perspective of representation learning \cite{golan2018deep,perera2020generative,xie2019spectral}, and the research on self-supervised learning is particularly bright. Contemporary there are two types of self-supervised learning methods as shown in Figure 1. Most GAN-based methods can be naturally regarded as belonging to generative self-supervised learning methods, such as OCGAN \cite{perera2019ocgan} and Old-Is-Gold \cite{zaheer2020old}. The generator tries to learn representations by reconstructing \emph{x} from the \emph{$x_{t}$} transformed by \emph{t}(\emph{x}), where \emph{t}(\emph{x}) is the image transformation determined by the pretext task. In order to avoid models focusing too much on pixel details, discriminative self-supervised learning methods for anomaly detection are gradually emerging. In these methods, representations are learned by predicting the labels ct provided by pretext task. For example, GEOM \cite{golan2018deep} performs efficient representation learning through the prediction of geometric transformations and improves the performance of anomaly detection as a pioneering work. After that, the potential of discriminative self-supervised learning in the field of anomaly detection began to show. For example, GOAD \cite{bergman2020classification} and SLOOD \cite{hendrycks2019using} detect abnormalities by multiple transformation classification. Furthermore, GDFR \cite{perera2020generative} and ARnet \cite{fei2020attribute} attempt to implement better representation learning by using generative method assisted discriminative method (called generative-discriminative representation learning). However, generator is difficult to learn abstract semantic representations from label prediction pretext tasks as effective as discriminator. Some samples produced by geometric transformation pretext tasks have the problem of label semantic ambiguity (shown in Figure 2), which means the generator cannot benefit from the pretext tasks designed for discriminative methods. On the other hand, discriminative methods mostly rely on the complex post-processing that require training sets in testing such as using Dirichlet score \cite{golan2018deep} and distribution normalization \cite{fei2020attribute}, which are high computation consumed.

In order to better representation learning for the generator, we try to combine generative methods with discriminative methods to mine as many meaningful representations as possible, and improve the anomaly detection performance of GAN in anomaly detection tasks. Figure 1(c) demonstrates the framework of our method. We reuse the discriminator as a predictor to guide the generator to generate samples that match the pretext task labels. The generator does not attempt to reconstruct the image on the pixel level, thus the representations learned do not focus too much on pixel details. On the other hand, the discriminator guides the generator to generate images of the correct categories and promotes the generator to learn accurate abstract semantic representations. Meanwhile, as other generative methods, we still use reconstruction error to represent the anomaly score, unlike some discriminative methods which require complicated post-processing.

Our method is different from the previous GAN-based approaches such as GDFR \cite{perera2020generative} and CompareGAN \cite{cvpr2019comparegan}. Their purpose is to make discriminator learn better representations assisted by generator, while our method is to make generator learn better representations guided by discriminator. We propose this discriminative-generative representation learning method for one-class anomaly detection task in this paper, named DGAD (discriminative-generative anomaly detection). The main contributions of us can be summarized as follows: 

1.The generator of DGAD can learn abstract semantic representation more efficiently without label semantic ambiguity problem. This is the first attempt to benefit generator from pretext tasks designed for discriminative methods as we know.

2.Our method significantly outperforms several state-of-the-arts on multiple benchmark datasets, increases the performance of the top-performing GAN-based baseline by 6\% on CIFAR-10 and 2\% on MVTAD.

3.Our experiments prove that based on the representations learned from the same pretext prediction task, the generative methods can approach the performance of discriminative methods by simply taking the reconstruction error as the anomaly score, and has a great advantage in speed.

4.Ablation studies show that absolute position information can degrade the representation learning ability of generative methods in geometric transformation pretext tasks, and has different effects on the representational learning ability of discriminative methods in different geometric transformation tasks, suggesting that the use of position information should be carefully selected according to different pretext tasks.

\section{Related Work}

\subsection{One-class anomaly detection (OCAD)}

OCAD assumes that all training samples belong to one class, and strives to learn a classification boundary that surrounds all normal samples. Thereby, any new sample that is not inside the decision boundary can be identified as an anomaly.

The performance of classical methods such as one-class SVM \cite{tax2004support} and one-class nearest neighbors \cite{tax2001uniform} usually limited by the dimensionality and complexity of the inputs, cannot be applied to high-dimensional or large scale datasets. In recent years, the development of deep learning improves the performance and practicability of OCAD \cite{oza2018one} \cite{abati2019latent}\cite{lee2018hierarchical}. An important kind of OCAD methods are based on auto-encoder \cite{nicolau2018learning}. The success of auto-encoder naturally attracts scholars to use GAN in order to obtain better detection performance.

\subsection{GAN-based anomaly detection}

According to the theory, the generated data distribution of GAN model will be consistent with the real data distribution under Nash equilibrium. Therefore, GAN can be used to detect anomalies outside the distribution. Based on this assumption, AnoGAN \cite{schlegl2017unsupervised} was proposed in 2017 with the use of GAN by the first time. However, the foundation of AnoGAN is based on intuition instead of theory. In fact, before AnoGAN, both BiGAN \cite{donahue2016adversarial} and ALI \cite{dumoulin2016adversarially} have already used GAN for adversarial learning and inference, which is the theoretical basis of applying GAN to representation learning. Then, CIAFL \cite{xie2017controllable} gives the research object of adversarial learning and inference: \emph{produce a data representation that maintains meaningful variations of data while eliminating noisy signals}. This definition exactly meets the requirements of anomaly detection. Thus, researchers started to take advantage of theories related with adversarial representation learning and inference to modify the GAN-based anomaly detection. For example, Efficient GAN-Anomaly \cite{zenati2018efficient} adopts the theory of BiGAN. ALAD \cite{zenati2018adversarially} adopts the theory of ALICE \cite{li2017alice}. GANomaly \cite{akcay2018ganomaly} further modifies the network structure and loss function to constrict latent space.

With the improvement of the GAN theory, GAN-based anomaly detection methods are divided into two directions. One is the deeper insight into latent space such as manifold learning \cite{pidhorskyi2018generative}, sparse regularization \cite{zhou2020sparse} and informative-negative mining \cite{perera2019ocgan}. The other is more accurate anomaly location, such as the use of class activation maps \cite{kimura2020adversarial} and attention expansion loss \cite{venkataramanan2020attention}. However, GAN-based representation learning has stagnated recently with only a few new research results \cite{wang2020transformation,zhai2019adversarial}. This situation limits the improvement of detection performance of GAN-based anomaly detection methods. Meanwhile, the rapid development of self-supervised learning provides some new inspiration for representation learning, promotes the use of pretext tasks in GAN-based anomaly detection method \cite{perera2020generative,bergman2020classification,hendrycks2019using,fei2020attribute}.

\begin{figure}[t]
\centerline{\includegraphics[width=0.5\textwidth]{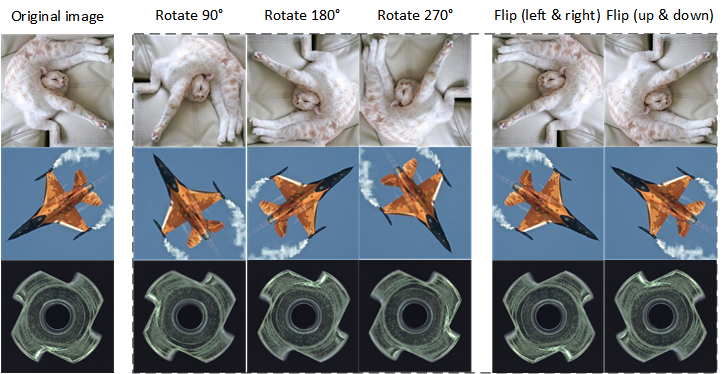}}
\caption{The example of label semantic ambiguity. Some images produced by the geometric transformations may have the same rationality as the original images, rendering the label meaningless.}
\label{Figure2}
\end{figure}

\subsection{Self-supervised learning}

As shown in Figure 1, contemporary self-supervised learning methods can roughly be broken down into two classes of methods in the field of computer vision: generative methods and discriminative methods. As a more traditional approach, generative methods focus on reconstruction error in the pixel space to learn representations, such as colorization \cite{zhang2016colorful}, super-resolution \cite{ledig2017photo}, inpainting \cite{pathak2016context}, and cross-channel prediction \cite{zhang2017split}. However, using pixel-level losses can lead to overly focus on pixel-level details, rather than more abstract latent representations, thereby reducing their ability to model correlations or complex structure. On the contrary, discriminative methods create (pseudo) labels by pretext tasks and learn representations by label predictions, such as image jigsaw puzzle \cite{noroozi2016unsupervised}, context prediction \cite{doersch2015unsupervised}, and geometric transformation recognition \cite{gidaris2018unsupervised}. Most of these pretext tasks have been used for anomaly detection in recent years \cite{golan2018deep,perera2020generative,bergman2020classification,hendrycks2019using,fei2020attribute}. In particular, as a kind of discriminative methods, contrastive learning methods treat each instance as a category, learn representations by contrasting positive and negative examples. Contrastive learning methods have led to great empirical success in computer vision tasks recently, such as MoCo \cite{he2020momentum} and SimCLR \cite{chen2020simple}. Some of techniques of contrast learning are just beginning to be used in anomaly detection \cite{tack2020csi,gong2019memorizing,park2020learning}. This shows that self-supervised learning has great potential in anomaly detection.

\begin{figure*}[ht]
\centerline{\includegraphics[width=1\textwidth]{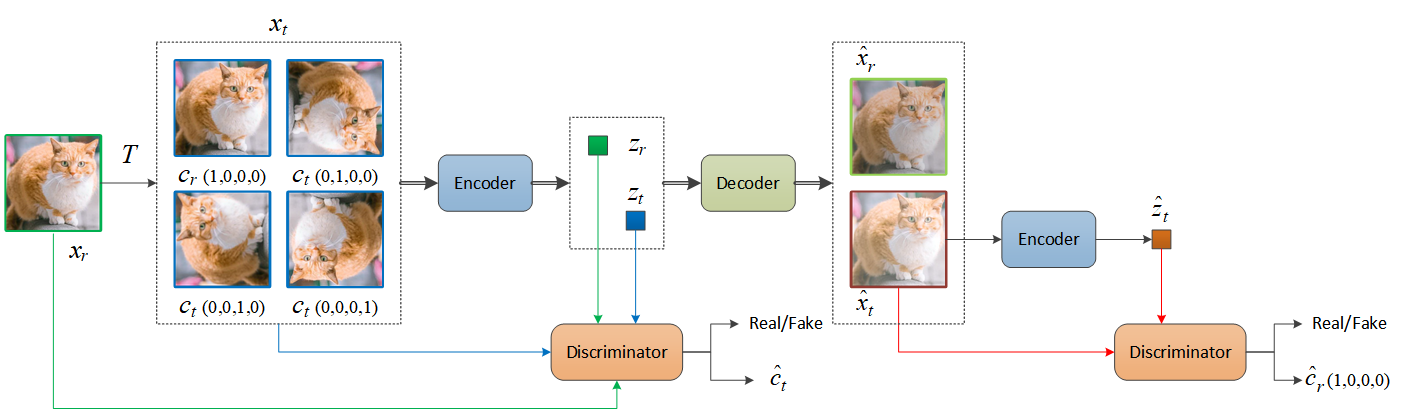}}
\caption{The training pipeline of DGAD with the pretext task of rotation prediction. For images $x_t$, the discriminator is trained as a predictor for predicting rotation angles $c_t$. For fake images $\hat{x}_t$, the parameters of the discriminator are fixed and expected to output a normal angle label $c_r$, thus guiding the decoder to generate a normal angle image.}
\label{Figure3}
\end{figure*}

\section{Our Method}

\subsection{Label semantic ambiguity}
As shown in Figure 2, some images produced by the geometric transformations are semantically indistinguishable from the original. The rotated image or flipped image may have the same rationality as the original image. In these cases, label prediction becomes meaningless since images with different labels all have the same semantic rationality. Most of the existing discriminative methods ignore this phenomenon because predictors are more or less tolerant to noise labels. Generative methods, however, are at a loss in the face of these pretext tasks designed for discriminative methods. Forced image generation based on meaningless labels will lead to confusion in model training and reduce the validity of representation. The problem of label semantic ambiguity prevents generators from learning representations from label prediction tasks. Hence, our method attempts to provide a solution that combines generative methods and discriminative methods to improve the representation learning ability of generative methods, while benefit from the pretext tasks designed for discriminative methods. 

\subsection{DGAD}
We propose a self-supervised learning method for one-class anomaly detection task based on generative adversarial nets, called DGAD. Like most GAN-based anomaly detection methods, our method is based on the assumption that the reconstruction error of the abnormal sample is greater than that of the normal sample. Our model uses pretext task of geometric transformations for representation learning. Take rotation prediction for example, as shown in Figure 3, we first rotate the training samples randomly and generate labels. Secondly, an encoder and a decoder are used to learn the encoding of samples in latent space by the image reconstruction of $x_r$, meanwhile a discriminator is used to predict the rotation angle of the image. Finally, the rotated image needs to be restored by the encoder, and the authenticity and rotation angle of the restored image are checked by the discriminator.

Mathematically, defining $\boldsymbol{X}$ to represent the domain of the data samples and $\boldsymbol{Z}$ to represent the domain of the encoding. Given a sample $x_r \in \boldsymbol{X}$ with label $c_r$ ($c_r$ is the same label that all the original samples have), we generate the transformed sample $x_t$ and transformation label $c_t$ by a transformer $\emph{T}$, then the encoder $\emph{En}$ converts them to encoded representation as $z_r$ and $z_t \in \boldsymbol{Z}$; and the decoder $\emph{De}$ is trained to reversely mapping them to $\hat{x}_r$ as follows
\begin{equation}
\emph{$x_{t}$}=T\left(\emph{$x_{r}$}\right)\label{eq1}
\end{equation}
\begin{equation}
\emph{$z_{r}$}=E n\left(\emph{$x_{r}$}\right), \emph{$z_{t}$}=E n\left(\emph{$x_{t}$}\right)\label{eq2}
\end{equation}
\begin{equation}
\hat{\emph{x}}_{r}=De\left(\emph{$z_{r}$}\right)=\hat{\emph{x}}_{t}=De\left(\emph{$z_{t}$}\right)\label{eq3}
\end{equation}

The discriminator \emph{D} is trained to distinguish between real and fake samples. Meanwhile, it contains a classifier \emph{$D_{cls}$} to predict the transformation label \emph{$c_{t}$} through the joint distributions of \emph{$x_{t}$} and \emph{$c_{t}$} as follows
\begin{equation}
c_{t}=\hat{c}_{t}=D_{c l s}\left(\emph{$x_{t}$}, \emph{$z_{t}$}\right)\label{eq4}
\end{equation}

Therefore, this trained classifier can be used to guide the decoder to generate the restored image with label \emph{$c_{r}$}.
\begin{equation}
c_{r}=\hat{c}_{r}=D_{c l s}\left(\hat{\emph{x}}_{t}, \hat{\emph{z}}_{t}\right)\label{eq5}
\end{equation}

Please note that unlike other generative methods, we do not train the decoder to restore \emph{$x_{r}$} from the transformed image \emph{$x_{t}$} by reconstruction loss, but by the label classified by $D_{cls}$. Images that cannot be predicted by classifiers need not be forced to be reconstructed, thus it avoids the meaningless reconstruction. Through this training strategy, our method both utilizes the representation learning ability of the generative methods for pixel details and the discriminative methods for abstract semantics. 

\subsection{Losses}

\textbf{Reconstruction Loss}. As a GAN-based method, it must have the reconstruction ability of normal images. We use the original image $\mathit{\boldsymbol{x}}_r$ and its reconstruction for the training of encoder \emph{En} and decoder \emph{De}.
\begin{equation}
L_{r e c}=\boldsymbol{E}_{\theta_{e n}, \theta_{d e}}\left\|\boldsymbol{x}_{r}-\hat{\boldsymbol{x}}_{r}\right\|_{1}\label{eq6}
\end{equation}

\noindent
where $\theta_{en}$ and $\theta_{de}$ denote the parameters of encoder and decoder respectively, $\|\cdot\|_{1}$ means L1 norm. The transformed images $\boldsymbol{x}_t$ are not included in this training, although we want $\hat{\boldsymbol{x}}_t$ to be the same as $\boldsymbol{x}_t$ after passing through the \emph{En} and \emph{De}. The restoration of $\boldsymbol{x}_t$ are guided by the discriminator.

\textbf{Classification Loss}. This objective has two terms: a loss of transformed images $\hat{\boldsymbol{x}}_t$ and their labels $\hat{\boldsymbol{c}}_t$ used to optimize the discriminator \emph{D}, and a loss of restored images used to optimize encoder and decoder. The former is defined as
\begin{equation}
L_{c l s}^{D}=\boldsymbol{E}_{\theta_{d}}\left[-\log D_{c l s}\left(\boldsymbol{c}_{t} \mid\left(\boldsymbol{x}_{t}, \boldsymbol{z}_{t}\right)\right)\right]\label{eq7}
\end{equation}

\noindent
where $\theta_{d}$ denotes the parameters of \emph{D}, $D_{cls}(\boldsymbol{c}|(\boldsymbol{x},\boldsymbol{z}))$ represents the joint probability distribution of $\boldsymbol{x}$ and $\boldsymbol{z}$ over labels $\boldsymbol{ c }$ computed by \emph{D}. \emph{D} learns how to classify transformed images by this loss.

The effective learning of the discriminator for classification does not mean that encoder and decoder also learn the relevant semantic representation. They need to learn effective representations at the same time, so as to help the discriminator classify more accurately. That requires encoder and decoder try to generate images that can be classified as the target label $\boldsymbol{c}_r$. Hence the loss used to optimize encoder and decoder is defined as
\begin{equation}
\begin{aligned}
L_{c l s}^{G} &=\boldsymbol{E}_{\theta_{e n}}\left[-\log D_{c l s}\left(\boldsymbol{c}_{t} \mid\left(\boldsymbol{x}_{t}, \boldsymbol{z}_{t}\right)\right)\right] \\
&+\boldsymbol{E}_{\theta_{e n}, \theta_{d e}}\left[-\log D_{c l s}\left(\boldsymbol{c}_{r} \mid\left(\hat{\boldsymbol{x}}_{t}, \hat{\boldsymbol{z}}_{t}\right)\right)\right]
\end{aligned}\label{eq8}
\end{equation}

The first term of this loss improves the efficiency of representation learning of encoder, and thus encoder can provide representation to the discriminator for more accurate classification. The latter term of this loss enables encoder and decoder to learn to restore the image to the correct category under the guidance of the discriminator. The discriminator first learns the image classification, then instructs encoder and decoder to generate normal category image. Image restoration is no longer guided by reconstruction loss but this loss. Thus, the representation of encoder and decoder learning is avoided to pay too much attention to pixel details; on the other hand, the interference of meaningless image reconstruction on representation learning is avoided.

\textbf{Compactness Loss}. The encoding of normal samples should be a compact distribution in latent space since images of the same category should be encoded close to each other. To do this, we constrain the variance among all encodings in the training batch. However, it is not reasonable to try to have each component in the encodings close together, which limits the inner-class diversity and causes the abnormal sample encoding to move closer to the normal sample encoding. Therefore, we only limit the variance on the channel axis. The compactness loss is defined as
\begin{equation}
L_{c m p}=\sqrt{\operatorname{\emph{$var_b$}}\left(\boldsymbol{z}_{r}\right)}\label{eq9}
\end{equation}

\noindent
where \emph{$var_b$}(·) means the variance among data from the same training batch,
\begin{equation}
\operatorname{\emph{$var_b$}}(\boldsymbol{z})=\boldsymbol{E}_{\theta_{e n}}\left(\operatorname{ap}_{c}\left(\emph{$z_{i}$}\right)-\frac{1}{N} \sum_{i=1}^{N} \operatorname{\emph{$ap_c$}}\left(\emph{$z_{i}$}\right)\right)^{2}\label{eq10}
\end{equation}

\noindent
where $N$ is the batch size, \emph{$ap_c$}(·) means the average pooling on the channel axis.

The encoding of all normal samples has a similar mean on each channel by this limitation, but can have different values on width axis and height axis. In other words, the channel means of encodings of the normal samples $\boldsymbol{z}_r$ are constrained in a compact high-dimensional distribution. The mean of the abnormal sample $\boldsymbol{z}_e$ on the channels will no longer be the same as the normal sample, resulting in significant differences between the $\boldsymbol{z}_r$ and $\boldsymbol{z}_e$, and helping to distinguish abnormal samples.

\textbf{Adversarial Loss}. We use the discriminator of the same architecture as BiGAN \cite{schlegl2017unsupervised} to train the model, and use hinge loss \cite{donahue2019large} to build the adversarial loss as follows
\begin{equation}
\begin{aligned}
L_{a d v}^{D} &=\boldsymbol{E}_{\theta_{d}}\left[\max \left(0,1-D_{a d v}\left(\boldsymbol{x}_{r}, \boldsymbol{z}_{r}\right)\right)\right] \\
&+\boldsymbol{E}_{\theta_{d}}\left[\max \left(0,1+D_{a d v}\left(\hat{\boldsymbol{x}}_{t}, \hat{\boldsymbol{z}}_{t}\right)\right)\right]
\end{aligned}\label{eq11}
\end{equation}
\begin{equation}
L_{a d v}^{G}=-\boldsymbol{E}_{\theta_{e n}, \theta_{d e}}\left[D_{a d v}\left(\hat{\boldsymbol{x}}_{t}, \hat{\boldsymbol{z}}_{t}\right)\right]\label{eq12}
\end{equation}

\noindent
where $D_{adv}(\boldsymbol{x},\boldsymbol{z})$ represents the discriminant value of joint probability distribution of $\boldsymbol{x}$ and $\boldsymbol{z}$ computed by \emph{D}.
Unlike receiving all transformed images in classification loss, the discriminator only considers $\boldsymbol{x}_r$ to be true images in adversarial loss, forcing the distribution of generated images to approximate the distribution of original images.

\textbf{Total Loss}. The objective to optimize \emph{En}, \emph{De} and \emph{D} are
\begin{equation}
L_{D}=L_{a d v}^{D}+\lambda_{c l s} L_{c l s}^{D}\label{eq13}
\end{equation}
\begin{equation}
L_{E n, D e}=L_{a d v}^{G}+\lambda_{c l s} L_{c l s}^{G}+\lambda_{r e c} L_{r e c}+\lambda_{c m p} L_{c m p}\label{eq14}
\end{equation}

\noindent
where $\lambda_{cls}$, $\lambda_{rec}$ and $\lambda_{cmp}$ are hyper-parameters. We use $\lambda_{cls}=10$, $\lambda_{rec}=20$ and $\lambda_{cmp}=100$ in all of our experiments.

\subsection{Pretext task and absolute position information}
In this section we describe the pretext tasks of geometric transformations we use in our study.

\textbf{Rotation}. An easy rotation mechanism that input images are rotated by {0°, 90°, 180°, 270°} is proposed by \cite{gidaris2018unsupervised}. In our study, we use a 4-bits one-hot encoding as the (pseudo) label of a rotated image. So all the original samples are set the same real label \emph{$c_{r}$} (1, 0, 0, 0) and the rotated image are the special label \emph{$c_{t}$} as shown in Figure 3. Unlike GEOM \cite{golan2018deep}, we don't add flips and shifts to the transformations since sometimes they cannot be classified meaningfully as we explained in Figure 2.

\textbf{Jigsaw Puzzle}. For centrally symmetric images (e.g. metal nuts and bottles in MVTec AD dataset), the pretext task of rotating or flipping the whole image is meaningless. Hence we apply Jigsaw Puzzle to transform training images. For learning the relative spatial position of feature in train data, each input image is split into $N$ $(N=4, 9, 16,..., n^2, n \in Z^+)$ partitions and each partition can be transformed independently. We set $\emph{N}=4$ and select to fixed the top-left partition and randomly permute other partitions with at least 2 displacements. The set of puzzles consist of 6 different permutations of the partitions in case of having 4 partitions. The split input image has 4 quadrants and each partition corresponds to each quadrant in original image.

There are three protocols in our study for one-class anomaly detection.

\textbf{Protocol 1}: Only using rotation on the original input images, every training samples is randomly rotated. So there are 4 different transformations and 4-bit one-hot label.

\textbf{Protocol 2}: Only using Jigsaw Puzzle on the split images. Randomly permuting the partition, and 6-bit one-hot (pseudo) labels are considered for the case of 6 possible transformations. In this protocol, the model need to learning the relative position of features.

\textbf{Protocol 3}: Based on Protocol 2, we select to add rotation on the partition after random permutation. So there are 6×64 different transformed possibilities. Considering that too many possibilities will increase the dimension of label, we suggest using multi-hot label instead of one-hot label in this protocol. In this protocol, the discriminator not only need to learn the relative position of features, but also the correct angle of each partition.

Since these geometric transformation pretext tasks are associated with location information, we explore the influence of absolute location information on the representational learning ability of our model We find the isolation of absolute position information can improve the performance of representation learning of generator, which will be proved in the ablation studies. Therefore, we use symmetry padding in accordance with reference \cite{islam2020much} as the default of our model.

\subsection{Anomaly score}

As a GAN-based framework, we define the anomaly score based on the reconstruction error. It contains two parts, one is the error between test image \emph{x} and the restored image $\hat{x}$, the other one is the error between \emph{z} and $\hat{z}$, correspond respectively the encoding of \emph{x} and $\hat{x}$. Anomaly score is the weighted sum of them as follows
\begin{equation}
s_{r e c}=\|\emph{x}-\hat{\emph{x}}\|_{1}+\lambda_{s}\|\emph{z}-\hat{\emph{z}}\|_{1}\label{eq15}
\end{equation}

\noindent
where $\lambda_{s}$=10 is the weight we used.

During test phase, the joint distribution of restored image and its encoding are mapped into the distribution of the training set. The in-distribution sample gets a lower anomaly score, otherwise high anomaly score for anomaly. At last, for the evaluation of area under the curve (AUC) of the receiver operating characteristic (ROC), we normalize the anomaly scores based on the test results as follows
\begin{equation}
s_{i}^{\prime}=\frac{s_{i}-\min \left(\boldsymbol{s}_{r e c}\right)}{\max \left(\boldsymbol{s}_{r e c}\right)-\min \left(\boldsymbol{s}_{r e c}\right)}\label{eq16}
\end{equation}

On the other hand, the trained discriminator enables us to use the predicted results for anomaly detection as well as other discriminative methods. Take Dirichlet score used in GEOM \cite{golan2018deep} as an example, the normality score of an image \emph{x} is
\begin{equation}
s_{d i r}=\sum_{i=0}^{k-1}\left(\alpha_{i}-1\right) \log D_{c l s}\left(T_{i}(\emph{x}), \emph{En}\left(T_{i}(\emph{x})\right)\right)\label{eq17}
\end{equation}

\noindent
where $T_i(x)$ is the $i^{th}$ transformation of $x$, $\alpha_{i}$ is a constant determined by $T_i(x)$. The Dirichlet score calculates the degree of anomaly of each sample. Higher scores indicate a more normal sample. We find that $s_{dir}$ can get better indicators than $s_{rec}$ in our experiment, however it needs to use the training data set and greatly increases the computational complexity. Nevertheless, we will present both scores in the ablation studies to provide options for trade-off between speed and performance.

\begin{table*}[ht]
\footnotesize
\centering
\caption{One-class anomaly detection AUC results for MNIST dataset.}
\setlength{\tabcolsep}{2.7mm}
{
\begin{tabular}{|l|c|c|c|c|c|c|c|c|c|c|c|}
\hline
\multicolumn{1}{|c|}{\textbf{Methods}} & \textbf{0} & \textbf{1} & \textbf{2} & \textbf{3} & \textbf{4}& \textbf{5} & \textbf{6}& \textbf{7} & \textbf{8} & \textbf{9} & \textbf{Mean} \\ \hline
\textbf{AnoGAN} \emph{IPMI’17}& 0.966& 0.992& 0.850& 0.887& 0.894 & 0.883& 0.947 & 0.935& 0.849& 0.924& 0.9127\\ \hline
\textbf{DSVDD} \emph{ICML’18} & 0.980& 0.997& 0.917& 0.919& 0.949 & 0.885& 0.983 & 0.946& 0.939& 0.965& 0.9480\\ \hline
\textbf{OCGAN} \emph{CVPR’19} & \textbf{0.998} & \textbf{0.999} & 0.942& 0.963& 0.975 & 0.980& 0.991 & 0.981& 0.939& 0.981& 0.9750\\ \hline
\textbf{LSA} \emph{CVPR’19} & 0.993& \textbf{0.999} & 0.959& 0.966& 0.956 & 0.964& 0.994 & 0.980& 0.953& 0.981& 0.9750\\ \hline
\textbf{CAVGA} \emph{ECCV’20} & 0.994& 0.997& \textbf{0.989} & \textbf{0.983} & 0.977 & 0.968& 0.988 & \textbf{0.986} & \textbf{0.988} & \textbf{0.991} & 0.9860\\ \hline
\textbf{DGAD} (Protocol 1)& 0.9934 & 0.9981 & 0.9876 & 0.9811 & \textbf{0.9853} & \textbf{0.988} & \textbf{0.9965} & 0.9837 & 0.9631 & 0.9872 & \textbf{0.9864} \\ \hline
\textbf{DGAD} (Protocol 2)& 0.9966 & 0.9987 & 0.9636 & 0.9549 & 0.9649& 0.9663 & 0.9915& 0.9668 & 0.9708 & 0.9735 & 0.9748\\ 
\hline
\end{tabular}}
\end{table*}

\begin{table*}[ht]
\footnotesize
\centering
\caption{One-class anomaly detection AUC results for CIFAR-10 dataset.}
\setlength{\tabcolsep}{3.2mm}
{
\begin{tabular}{|l|c|c|c|c|c|c|c|c|c|c|c|}
\hline
\multicolumn{1}{|c|}{\textbf{Methods}} & \textbf{Plane} & \textbf{Car} & \textbf{Bird}& \textbf{Cat} & \textbf{Deer}& \textbf{Dog} & \textbf{Frog}& \textbf{Horse} & \textbf{Ship}& \textbf{Truck} & \textbf{Mean} \\ \hline
\textbf{AnoGAN} \emph{IPMI’17}& 0.671& 0.547& 0.529& 0.545& 0.651& 0.603& 0.585& 0.625& 0.758& 0.665& 0.6179\\ \hline
\textbf{LSA} \emph{CVPR’19} & 0.735& 0.580 & 0.690 & 0.542& 0.761& 0.546& 0.751& 0.535& 0.717& 0.548& 0.641 \\ \hline
\textbf{DSVDD} \emph{ICML’18} & 0.617 & 0.659 & 0.508& 0.591& 0.609& 0.657& 0.677& 0.673& 0.759& 0.731& 0.6481\\ \hline
\textbf{OCGAN} \emph{CVPR’19} & 0.757& 0.531 & 0.640 & 0.620 & 0.723& 0.620 & 0.723& 0.575& 0.820 & 0.554& 0.6566\\ \hline
\textbf{CAVGA} \emph{ECCV’20} & 0.653& 0.784& \textbf{0.761} & \textbf{0.747} & \textbf{0.775} & 0.552& \textbf{0.813} & 0.745 & 0.801 & 0.741 & 0.737 \\ \hline
\textbf{DROCC} \emph{ICML’20} & \textbf{0.817} & 0.767& 0.667& 0.671& 0.736 & 0.744 & 0.744 & 0.714& 0.800& 0.762& 0.7423 \\ \hline
\textbf{DGAD} (Protocol 1)& 0.746& \textbf{0.876} & 0.732& 0.711& 0.766& \textbf{0.804} & 0.760 & \textbf{0.884} & \textbf{0.862} & \textbf{0.872} & \textbf{0.8012} \\ \hline
\textbf{DGAD} (Protocol 2)& 0.800& 0.855& 0.688& 0.658& 0.662& 0.761& 0.713& 0.805& 0.854& 0.833& 0.7629\\ \hline
\end{tabular}}
\end{table*}

\section{Experiments}

\subsection{Datasets and baselines}
We evaluated our model on MNIST \cite{lecun1998mnist}, CIFAR-10 \cite{krizhevsky2009learning}, and MVTAD \cite{bergmann2019mvtec} for one-class anomaly detection. We compare our method to several classical methods and state-of-the-art GAN-based methods for one-class anomaly detection. They are AnoGAN \cite{schlegl2017unsupervised}, DSVDD \cite{ruff2018deep}, GEOM \cite{golan2018deep}, OCGAN \cite{perera2019ocgan}, LSA \cite{abati2019latent}, DROCC \cite{goyal2020drocc} and CAVGA \cite{venkataramanan2020attention}. Most of the results of them are the reported results in the original papers. 

\subsection{Network structure and training details}
The encoder consists of several convolution layers and residual blocks \cite{he2016deep}. The input enters a 7×7 convolution layer with stride 1, two 4×4 convolution layers with stride 2 for down-sampling, three residual blocks layers and a 3×3 convolution layer with stride 1 in turn. Instance normalization (IN) \cite{ulyanov2016instance} is used in all layers follows by a ReLU activation except the last output layer. A tanh activation was placed after the last convolution layer to restrict the output of the latent dimension. The decoder is the symmetry of the encoder. The difference is that following by a bilinear interpolation and 5×5 convolution layer for up-sampling after three residual blocks. The number of parameters of encoder and encoder is about 8.9M.

The discriminator is based on the BiGAN architecture. After two convolution layers for down-sampling, $z$ concatenates with the feature map of $x$ and into the next convolution layer. Leaky ReLU is used in all layers with a negative slope of 0.01 except the last output layer. The number of parameters of discriminator is about 3.5M.

Our model is trained using Adam with $\beta_1$ = 0.5 and $\beta_2$ = 0.999. The batch size is set to 64 for CIFAR-10 and MNIST dataset. We perform one by one update for generator and discriminator with an initial learning rate of 0.0001 for 10000 iterations. We use spectral normalization \cite{miyato2018spectral} in the discriminator. We use symmetry padding as defaults in convolution. The model is implemented by Tensorflow 2.3$\footnote{Our code and models are available at github.com/*.}$, and the training takes about 50 minutes per class on a single NVIDIA 2080TI GPU for CIFAR-10. 

\subsection{MNIST and CIFAR-10 for anomaly detection}

In order to achieve one-class detection, one class at each time is considered as the normal class in training, and other class are regarded as anomaly in testing. We evaluate the performance by AUC which is commonly used for evaluating performance in anomaly detection tasks. The results of anomaly detection on MNIST and CIFAR-10 are presented in Table 1 and Table 2 respectively. They show the AUC value of each class and the total average AUC values. The proposed DGAD with \textbf{Protocol 1} outperforms the compared methods for MNIST. DGAD with \textbf{Protocol 1} and \textbf{Protocol 2} both outperform the compared methods on CIFAR-10. DGAD increases the performance of the top-performing GAN-based baseline (CAVGA) by 6\% on CIFAR-10.

However, the performance of \textbf{Protocol 2} is weaker than that of \textbf{Protocol 1}, indicating that rotation is a more effective pretext task for our method. And the simplicity of rotation makes its advantages more obvious. However, the rotation-based pretext task is meaningless for a centrally-symmetric image, which limits its application. The existence of \textbf{Protocol 2} and \textbf{Protocol 3} is still necessary.

\begin{table*}[ht]
\footnotesize
\centering
\caption{One-class anomaly detection AUC results for MVTec AD dataset.}
{
\begin{tabular}{|l|c|c|c|c|c|c|c|c|c|c|c|c|c|c|c|c|}
\hline
\multicolumn{1}{|c|}{\textbf{Methods}} & \textbf{0}& \textbf{1}& \textbf{2}& \textbf{3}& \textbf{4}& \textbf{5}& \textbf{6}& \textbf{7}& \textbf{8}& \textbf{9}& \textbf{10} & \textbf{11} & \textbf{12} & \textbf{13} & \textbf{14} & \textbf{Mean}\\ \hline
\textbf{AnoGAN} \emph{IPMI’17}& 0.69& 0.50 & 0.58& 0.50 & 0.52& 0.62& 0.68& 0.49& 0.51& 0.51& 0.53& 0.67& 0.57& 0.35& 0.59& 0.55 \\ \hline
\textbf{LSA} \emph{CVPR’19} & 0.86& 0.80 & 0.71& 0.67& 0.70 & 0.85& 0.75& \textbf{0.74} & 0.70 & 0.54& 0.61& 0.50 & 0.89& 0.75& \textbf{0.88} & 0.73 \\ \hline
\textbf{CAVGA} \emph{ECCV’20} & 0.89& \textbf{0.84} & \textbf{0.83} & 0.67& 0.71& \textbf{0.88} & \textbf{0.85}& 0.73& 0.70 & \textbf{0.75} & 0.63& 0.73& 0.91& \textbf{0.77} & 0.87& 0.78 \\ \hline
\textbf{DGAD} (Protocol 3)& \textbf{0.97} & 0.80& 0.60& \textbf{0.95} & \textbf{0.94} & 0.76& 0.72 & 0.52& \textbf{0.83} & 0.67& \textbf{0.90} & \textbf{0.88} & \textbf{0.93} & 0.67& 0.82& \textbf{0.80} \\ \hline
\end{tabular}}
\par{\leftline{\small *0$\sim$14 denote bottle, hazelnut, capsule, metal nut, leather, pill, wood, carpet, tile, grid, cable, transistor, toothbrush, screw, zipper.}}
\end{table*}

\begin{table*}[ht]
\footnotesize
\centering
\caption{Ablation studies and comparison experiments on CIFAR-10 dataset.}
{
\begin{tabular}{|c|lcccccccccc|c|}
\hline
\textbf{}& \multicolumn{1}{c|}{\textbf{Methods}} & \multicolumn{1}{c|}{\textbf{Plane}} & \multicolumn{1}{c|}{\textbf{Car}} & \multicolumn{1}{c|}{\textbf{Bird}} & \multicolumn{1}{c|}{\textbf{Cat}} & \multicolumn{1}{c|}{\textbf{Deer}}& \multicolumn{1}{c|}{\textbf{Dog}} & \multicolumn{1}{c|}{\textbf{Frog}}& \multicolumn{1}{c|}{\textbf{Horse}} & \multicolumn{1}{c|}{\textbf{Ship}}& \textbf{Truck} & \textbf{Mean} \\ 
\hline
\textbf{Protocol 1} & \multicolumn{1}{l|}{\textbf{GAN}} & {\textbf{0.794}} & 0.733 & 0.708 & 0.633 & 0.723 & 0.705 & 0.706 & 0.702 & 0.859 & 0.701 & 0.7265\\
& \multicolumn{1}{l|}{\textbf{DGAD - CL}} & 0.735 & 0.863 & 0.709 & 0.666 & 0.698 & 0.766 & 0.726 & 0.868 & 0.858 & 0.832 & 0.7721\\
& \multicolumn{1}{l|}{\textbf{DGAD + zero-padding}} & 0.743& 0.855& 0.692& 0.705 & 0.758& \textbf{0.819}& 0.736 & 0.865 & 0.854& 0.847& 0.7874\\
& \multicolumn{1}{l|}{\textbf{DGAD + coord}}& 0.767& 0.853 & 0.692 & 0.676& 0.750& 0.763& \textbf{0.798}& 0.813& 0.848 & 0.830& 0.7797 \\
& \multicolumn{1}{l|}{\textbf{DGAD}}& 0.746 & \textbf{0.876}& \textbf{0.732} & \textbf{0.711}& \textbf{0.766}& 0.804 & 0.760& \textbf{0.884}& \textbf{0.862}& \textbf{0.872} & \textbf{0.8012} \\ \cline{2-13} 
& \multicolumn{1}{l|}{\textbf{DGAD + zero-padding (\emph{s}$_{dir}$)}} & {0.746}& {\textbf{0.915}} & {0.752} & {0.699}& {0.781}& {0.810} & {\textbf{0.799}} & {0.920} & {\textbf{0.904}} & {\textbf{0.887}} & \textbf{0.8212} \\
& \multicolumn{1}{l|}{\textbf{DGAD + coord (\emph{s}$_{dir}$)}} & 0.746 & 0.914 & \textbf{0.755} & 0.672 & \textbf{0.794}& \textbf{0.811}& 0.798 & \textbf{0.929}& 0.895 & {0.874}& 0.8207\\
& \multicolumn{1}{l|}{\textbf{DGAD (\emph{s}$_{dir}$)}} & \textbf{0.752}& 0.910& 0.730 & \textbf{0.700}& 0.785 & 0.810& 0.763 & 0.915 & 0.901 & {0.878}& 0.8144\\ \hline
\textbf{Protocol 2} & \multicolumn{1}{l|}{\textbf{GAN}} & {0.772}& {0.803}& {0.615} & {0.621}& {\textbf{0.683}} & {0.648}& {0.684}& {0.665}& {0.762}& {0.833}& 0.7086\\
& \multicolumn{1}{l|}{\textbf{DGAD - CL}} & 0.769 & 0.823 & 0.658& \textbf{0.690} & 0.671 & 0.702 & 0.701 & 0.793 & 0.813 & {0.802}& 0.7422\\ 
& \multicolumn{1}{l|}{\textbf{DGAD + zero-padding}} & 0.687 &0.819	&0.600	&0.655	&0.583	&0.738	&0.699	&0.708	&0.844& {0.764}& 0.7096\\
& \multicolumn{1}{l|}{\textbf{DGAD + coord}}& 0.713	&0.827	&0.536	&0.672	&0.631	&0.708	&0.659	&0.771	&\textbf{0.862}	&0.816	&0.7195  \\
& \multicolumn{1}{l|}{\textbf{DGAD}}& \textbf{0.800}& \textbf{0.855}& \textbf{0.688} & 0.658 & 0.662 & \textbf{0.761}& \textbf{0.713}& \textbf{0.805}& 0.854& {\textbf{0.833}} & \textbf{0.7629} \\ \cline{2-13} 
& \multicolumn{1}{l|}{\textbf{DGAD + zero-padding (\emph{s}$_{dir}$)}} & 0.718	&\textbf{0.834}	&0.581	&0.609	&0.609	&0.675	&0.717	&0.772	&0.796	&0.806	&0.7115  \\
& \multicolumn{1}{l|}{\textbf{DGAD + coord (\emph{s}$_{dir}$)}} & 0.710	&0.820	&0.548	&0.559	&0.563	&0.629	&0.603	&0.756	&0.812	&0.802	&0.6802 \\
& \multicolumn{1}{l|}{\textbf{DGAD (\emph{s}$_{dir}$)}}& \textbf{0.787}& {0.825}& \textbf{0.721} & \textbf{0.695}& \textbf{0.755}& \textbf{0.778}& \textbf{0.736}& \textbf{0.815}& \textbf{0.872}& \textbf{0.857}& \textbf{0.7840} \\ \hline
\end{tabular}}
\end{table*}

\subsection{MVTec for anomaly detection}

For MVTec AD dataset, we perform random zoom augmentation for each category and resize all the image to 128×128 in training and testing. Training is conducted for 6000 epochs with batch size 8 on normal data. 

Considering that many images in the data set are either centrally symmetric or textured, we use \textbf{Protocol 3} as the pretext task. As shown in Table 3, our method achieves the best average performance (2\% higher than CAVGA). However, we can see from the table that our method is not suitable for detecting texture anomalies (e.g. carpet and grid), and some tiny anomalies (e.g. capsule and screw). This is the inherent defect of relying on the reconstruction error to measure the anomaly, which will be the focus of improvement in future work.

\subsection{Ablation studies}
The ablation studies have three purposes: verify the correctness of each conjecture in our method, compare the differences between generative method and discriminative method, and explore the influence of absolute location information on self-supervised representation learning.

All the studies are on the CIFAR-10 dataset as Table 4 shows. We compared \textbf{Protocol 1} with \textbf{Protocol 2} based on DGAD. There are two parts ($s_{rec}$ and $s_{dir}$) and eight variants both in \textbf{Protocol 1} and \textbf{Protocol 2}. \textbf{GAN} represents learning representations only by pixel-level image reconstruction, not by discriminator guidance. \textbf{DGAD - CL} represents the DGAD without compactness loss. \textbf{DGAD + zero-padding} represents using zero padding instead of symmetry padding in the convolution (introducing absolute location information). \textbf{DGAD + coord} represents using coordconv \cite{liu2018intriguing} for more direct absolute location information. $\boldsymbol{(s_{dir})}$ represents to calculate the Dirichlet score $s_{dir}$ using only the discriminator.

As we can see in Table 4, DGAD with \textbf{Protocol 1} and \textbf{Protocol 2} both significantly increase AUC by 5.4$\sim$7.5\% compared to \textbf{GAN}, which proves that the guidance of discriminator is more beneficial to representation learning than pixel-level reconstruction. And the absence of compactness loss reduces performance by 2$\sim$2.9\%, which demonstrates its effectiveness.

In part one of \textbf{Protocol 1}, the experimental results of zero-padding and coordconv confirm the interference of absolute position information on representation learning. With the help of absolute location information for geometric transformation prediction, encoder and decoder become lazy to learn more semantic representations, resulting in a 1$\sim$2\% performance reduction.

In part two of \textbf{Protocol 1}, on the contrary, the discriminator improves the performance of anomaly detection with the aid of absolute position information, which indicates that the absolute location information can make it easier for the discriminator to classify and recognize objects and scenes, but it will correspondingly weaken the generator's learning of abstract representations.

However, the influence of absolute location information in \textbf{Protocol 2} is changed. The performance of both generator and discriminator is degraded with absolute location information. Experiments show that the effect of location information on performance is related to the choice of pretext tasks. Further experiments show that this is caused by the discriminator's generalization ability under different pretext tasks (see supplementary materials).  Therefore, we suggest making a careful analysis of the tasks to be faced before model design.

\begin{table}[t]
\footnotesize
\centering
\caption{Speed comparison on CIFAR-10 between \emph{{s}$_{rec}$} and \emph{{s}$_{dir}$}.}
\begin{tabular}{|c|c|c|c|}
\hline
\multicolumn{2}{|c|}{\textbf{Protocol 1}} & \multicolumn{2}{c|}{\textbf{Protocol 2}} \\ \hline
DGAD($s_{rec}$) & DGAD($s_{dir}$)&DGAD($s_{rec}$)& DGAD($s_{dir}$)\\ \hline
\small
120.97 im/s & 33.04 im/s& 121.66 im/s& 14.18 im/s\\ \hline
\end{tabular}
\par{\leftline{\small *im/s means images per second.}}
\end{table}

It is worth noting that $s_{rec}$ is slightly lower than $s_{dir}$ by 1$\sim$3\%. This shows that although the reconstruction loss is simple, it can be close to the score of discriminative method which relies on complex post-processing. And our approach shows a great advantage in speed as shown in Table 5. The calculation of $s_{rec}$ is 3.6 to 8.5 times faster than that of $s_{dir}$, even though the discriminator has a smaller number of parameters than the generator. The more kinds of transformation, the more time it takes to calculate $s_{dir}$, while the calculation of $s_{rec}$ has nothing to do with the pretext task and has a constant computation.

\section{CONCLUSION}

In this paper, we propose a novel GAN-based anomaly detection method with the help of discriminative-generative representation learning, named DGAD. Under the guidance of discriminator, the generator can better learn representation and avoid the problem of label semantic ambiguity. DGAD increases the performance of the top-performing GAN-based baseline by 6\% on CIFAR-10 and 2\% on MVTAD. DGAD approaches the performance of discriminative method and has a great advantage in speed. As an additional conclusion, we find that absolute position information has different effects on the representational learning ability of generative methods and discriminative methods, which indicates that the use of absolute position information should be carefully selected according to different pretext tasks. Considering the natural relationship between various pretext tasks and absolute location information, the interaction between them should arouse more attention of researchers. We believe that the representations learned by the generator should be better quantified to accurately measure the degree of abnormality, which will be the focus of our future research.

{\small
\bibliographystyle{ieee_fullname}
\bibliography{egbib}
}

\clearpage
\section{Supplementary Materials}
In this document, we provide more experimental results and analyses that have not been presented in the original due to the space limitations.

\subsection{Qualitative analysis of model performance}
This section consists of two purposes: one is to verify that the encoder and decoder have indeed learned the abstract semantic representation; the other is to visualize the test results to verify our theory.

\begin{figure}[ht]
\centerline{\includegraphics[width=0.45\textwidth]{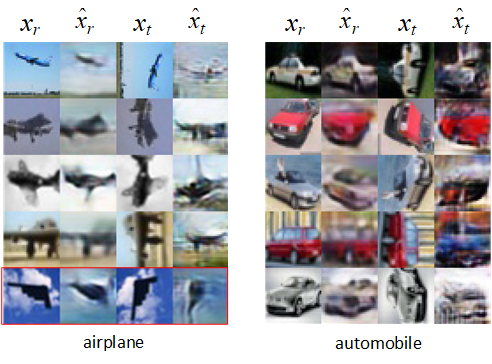}}
\caption{The reconstruction results of the rotated images by encoder and decoder.}
\label{Figure4}
\end{figure}

Figure 4 shows the reconstruction results of the rotated images by encoder and decoder. The figure shows the results of two models, one for airplane and the other for automobile. As we can see, our model can reconstruct the original image $\boldsymbol{x}_{r}$ and restore the rotated image $\boldsymbol{x}_{t}$ to the normal angle image $\hat{x}_{t}$. This confirms that our model achieves the goal of the pretext task. On the other hand, we can see that the restored image does not always match the original image, because we do not impose pixel-level consistency constraints. This confirms that our model indeed learned abstract semantic representations of geometric features, not just textures. In particular, the red box shows an example of a rotated image that remains unchanged before and after restoration. This phenomenon is consistent with our hypothesis. Since the rotated image still retains the visual rationality, the discriminator cannot predict its rotation angle. Therefore, the decoder only reconstructs the image without forcibly restoring it to the original image. Our model has better representation learning ability because it avoids the meaningless image restoration.

\begin{figure}[ht]
\centerline{\includegraphics[width=0.28\textwidth]{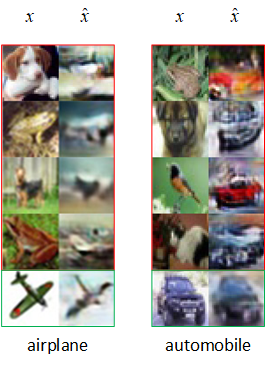}}
\caption{The reconstruction results of test images.}
\label{Figure5}
\end{figure}

Figure 5 shows the reconstruction results of test images. We can see that the reconstruction results of out-of-distribution images tend to be in-distribution images, resulting in a huge difference from the original images. Meanwhile, the normal images can be reconstructed into similar images, which provides us with a tool to distinguish the anomaly by the reconstruction error.

\subsection{Quantitative analysis of model performance}
This section consists of two purposes: the first is to verify the authenticity of our experimental results through visualization, and the second is to verify the relevant conclusions in the paper through supplementary experiment results.

\begin{figure}[ht]
\centering
\subfigure[The distribution of anomaly scores.]{
\begin{minipage}[b]{0.4\textwidth}
\includegraphics[width=1\textwidth]{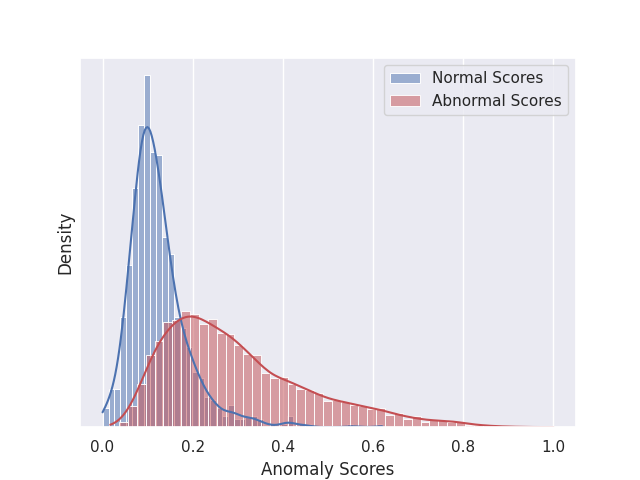}
\end{minipage}
}
\subfigure[The ROC curve of detection.]{
\begin{minipage}[b]{0.4\textwidth}
\includegraphics[width=1\textwidth]{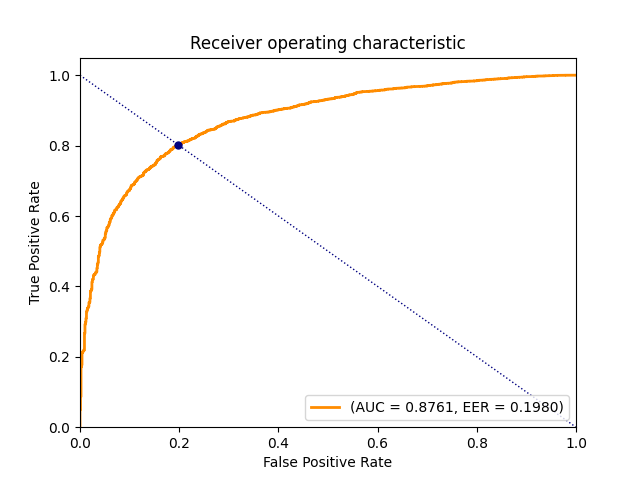}
\end{minipage}
}
\centering
\caption{The visualization of anomaly detection results of automobile.}
\vspace{0.2in}
\end{figure}

Figure 6 shows the visualization of anomaly detection results of automobile. As can be seen from the figure, there are two different distributions of anomaly scores of abnormal and normal in Figure 6(a). The anomaly scores of normal images are concentrated in low scores, while those of abnormal images are the opposite. However, the two distributions still have overlapping areas, leading to an error rate of 0.2 (Figure 6(b)), which needs to be improved in the future.

As the ablation studies and comparison experiments on CIFAR-10 dataset shown in Table 4 in the paper, absolute position information can degrade the representation learning ability of generator but benefit discriminator in \textbf{Protocol 1}. However, the Dirichlet scores of \textbf{DGAD + zero-padding (\emph{{s}$_{dir}$})} and \textbf{DGAD + coord (\emph{{s}$_{dir}$})} are also degraded in \textbf{Protocol 2}, even worse. To study the causes of this phenomenon, we compare the classification accuracy of discriminator as shown in Table 6 and Table 7. 

\begin{table*}[ht]
\footnotesize
\centering
\caption{Classification accuracy of discriminator (only normal samples in test set).}
{
\begin{tabular}{|l|lcccccccccc|c|}
\hline
\multicolumn{1}{|c|}{\textbf{}}      & \multicolumn{1}{c|}{\textbf{Methods}} & \multicolumn{1}{c|}{\textbf{Plane}} & \multicolumn{1}{c|}{\textbf{Car}} & \multicolumn{1}{c|}{\textbf{Bird}} & \multicolumn{1}{c|}{\textbf{Cat}} & \multicolumn{1}{c|}{\textbf{Deer}} & \multicolumn{1}{c|}{\textbf{Dog}} & \multicolumn{1}{c|}{\textbf{Frog}} & \multicolumn{1}{c|}{\textbf{Horse}} & \multicolumn{1}{c|}{\textbf{Ship}} & \textbf{Truck} & \textbf{Mean}  \\ \hline
\textbf{Protocol 2} & \multicolumn{1}{l|}{\textbf{DGAD + zero-padding}}       & 0.946                               & \textbf{0.979}                    & \textbf{0.961}                     & 0.959                             & \textbf{0.966}                     & 0.966                             & \textbf{0.973}                     & \textbf{0.971}                      & 0.962                              & \multicolumn{1}{c}{\textbf{0.971}} & \textbf{0.965} \\ 
                                     & \multicolumn{1}{l|}{\textbf{DGAD +   coord}}            & \textbf{0.957}                      & \multicolumn{1}{c}{0.976}                             & 0.943                              & \textbf{0.971}                    & 0.941                              & \textbf{0.974}                    & 0.966                              & 0.968                               & \textbf{0.967}                     & \multicolumn{1}{c}{0.965}          & 0.963          \\ 
                                     & \multicolumn{1}{l|}{\textbf{DGAD}} & 0.920          & 0.969        & 0.906         & 0.916        & 0.920         & 0.937        & 0.935         & 0.941          & 0.946         & \multicolumn{1}{c}{0.946}          & 0.934          \\ \hline
\end{tabular}}
\end{table*}

\begin{table*}[ht]
\footnotesize
\centering
\caption{Classification accuracy of discriminator (all samples in test set).}
{
\begin{tabular}{|l|lcccccccccc|c|}
\hline
\multicolumn{1}{|c|}{\textbf{}}      & \multicolumn{1}{c|}{\textbf{Methods}} & \multicolumn{1}{c|}{\textbf{Plane}} & \multicolumn{1}{c|}{\textbf{Car}}   & \multicolumn{1}{c|}{\textbf{Bird}}  & \multicolumn{1}{c|}{\textbf{Cat}}   & \multicolumn{1}{c|}{\textbf{Deer}}  & \multicolumn{1}{c|}{\textbf{Dog}}   & \multicolumn{1}{c|}{\textbf{Frog}}  & \multicolumn{1}{c|}{\textbf{Horse}} & \multicolumn{1}{c|}{\textbf{Ship}}  & \textbf{Truck} & \textbf{Mean}  \\ \hline
\textbf{Protocol 2} & \multicolumn{1}{l|}{\textbf{DGAD + zero-padding}}       & 0.882                               & 0.840                      & 0.911                      & 0.942                               & 0.881                      & 0.925                               & 0.905                      & 0.890                      & 0.848                               & \multicolumn{1}{c}{0.869} & 0.889 \\  
                                     & \multicolumn{1}{l|}{\textbf{DGAD +   coord}}            & 0.854                      & 0.832                               & 0.928                               & 0.929                      & 0.905                               & 0.925                      & 0.904                               & 0.889                               & 0.830                      & \multicolumn{1}{c}{0.870}          & 0.887          \\  
                                     & \multicolumn{1}{l|}{\textbf{DGAD}} & \textbf{0.409} & \textbf{0.507} & \textbf{0.433} & \textbf{0.355} & \textbf{0.254} & \textbf{0.516} & \textbf{0.397} & \textbf{0.405} & \textbf{0.605} & \multicolumn{1}{c}{\textbf{0.768}} & \textbf{0.465} \\ \hline
\end{tabular}}
\end{table*}

We only use the normal samples in the test set to calculate the prediction accuracy of the discriminator in Table 6. For each image, we generate six kinds of transformed images and their corresponding labels according to the pretext task. We can see that absolute position information does improve the accuracy of the prediction by about 3\%. However, when we use the whole test set to calculate the prediction accuracy as show in Table 7, we can find that the use of absolute position information greatly improves the prediction accuracy for abnormal samples. The absolute position information enables the discriminator to obtain good prediction accuracy for both normal samples and abnormal samples (more than 0.88), which leads to the reduction of the difference between their abnormal scores. On the contrary, \textbf{DGAD} without absolute position information has poor prediction accuracy for abnormal samples (lower than 0.47), which makes normal and abnormal can be well distinguished. The absolute position information makes the discriminator too generalizable, which is the reason for the performance degradation of anomaly detection.

Interestingly, we did not find this phenomenon in the rotation pretext task, which supports the point in our paper: the use of absolute position information should be carefully selected according to different pretext tasks.

\subsection{Code}
We provide the code of DGAD based on CIFAR-10. Reviewers can use our code to reproduce or test our model.

Reviewers may need to upgrade Seaborn and Matplot to run the tests correctly.

Train: {\tt python main.py}

This command will train each of the 10 categories of CIFAR-10 in turn.

Test: {\tt python main.py --phase test}

This command will test each of the 10 categories of CIFAR-10 in turn. Or,

Test: {\tt python main.py --phase test --test\_object 1}

This command will test category 1 (automobile) of CIFAR-10.

The visual results (distribution diagram and ROC) of the test can be found in {\tt ./results}.

\end{document}